\documentclass[final]{cvpr}

\usepackage{times}
\usepackage{epsfig}
\usepackage{graphicx}
\usepackage{booktabs} 
\usepackage{amsmath}
\usepackage{amssymb}
\usepackage{algorithm}
\usepackage{algorithmic}
\usepackage{multirow}
\usepackage{bbm}
\usepackage[table,xcdraw]{xcolor}
\usepackage{tabu}

\usepackage[pagebackref=true,breaklinks=true,colorlinks,bookmarks=false]{hyperref}

\begin{document}

\title{BiOpt: Bi-Level Optimization for Few-Shot Segmentation}

\author{Jinlu Liu, Liang Song, Yongqiang Qin\\
AInnovation Tecnology Co., Ltd.\\
{\tt\small liujinlu, songliang, qinyongqiang@ainnovation.com}}

\maketitle

\begin{abstract}
Few-shot segmentation is a challenging task that aims to segment objects of new classes given scarce support images. In the inductive setting, existing prototype-based methods focus on extracting prototypes from the support images; however, they fail to utilize semantic information of the query images. In this paper, we propose Bi-level Optimization (BiOpt), which succeeds to compute class prototypes from the query images under inductive setting. The learning procedure of BiOpt is decomposed into two nested loops: inner and outer loop. On each task, the inner loop aims to learn optimized prototypes from the query images. An init step is conducted to fully exploit knowledge from both support and query features, so as to give reasonable initialized prototypes into the inner loop. The outer loop aims to learn a discriminative embedding space across different tasks. Extensive experiments on two benchmarks verify the superiority of our proposed BiOpt algorithm. In particular, we consistently achieve the state-of-the-art performance on 5-shot PASCAL-$5^i$ and 1-shot COCO-$20^i$.
\end{abstract}

\section{Introduction}
Remarkable progress has been achieved in various segmentation tasks ranging from semantic segmentation \cite{long2015fully, chen2017deeplab, chen2018encoder} to panoptic segmentation \cite{kirillov2019panoptic, cheng2020panoptic}. These achievements normally take advantage of large-scale dataset that is costly and difficult to obtain, especially in real-world scenarios. In conventional semantic segmentation, abundant pixel-wise annotations are required to train a good segmentation model. Much efforts have been made to reduce the need of annotation in the segmentation community. Recently, few-shot segmentation arouses much attention which aims to transfer knowledge learnt on base class to segment objects of novel class given few support images.

\begin{figure}
\centering
\includegraphics[width=3.3in, height=1.15in]{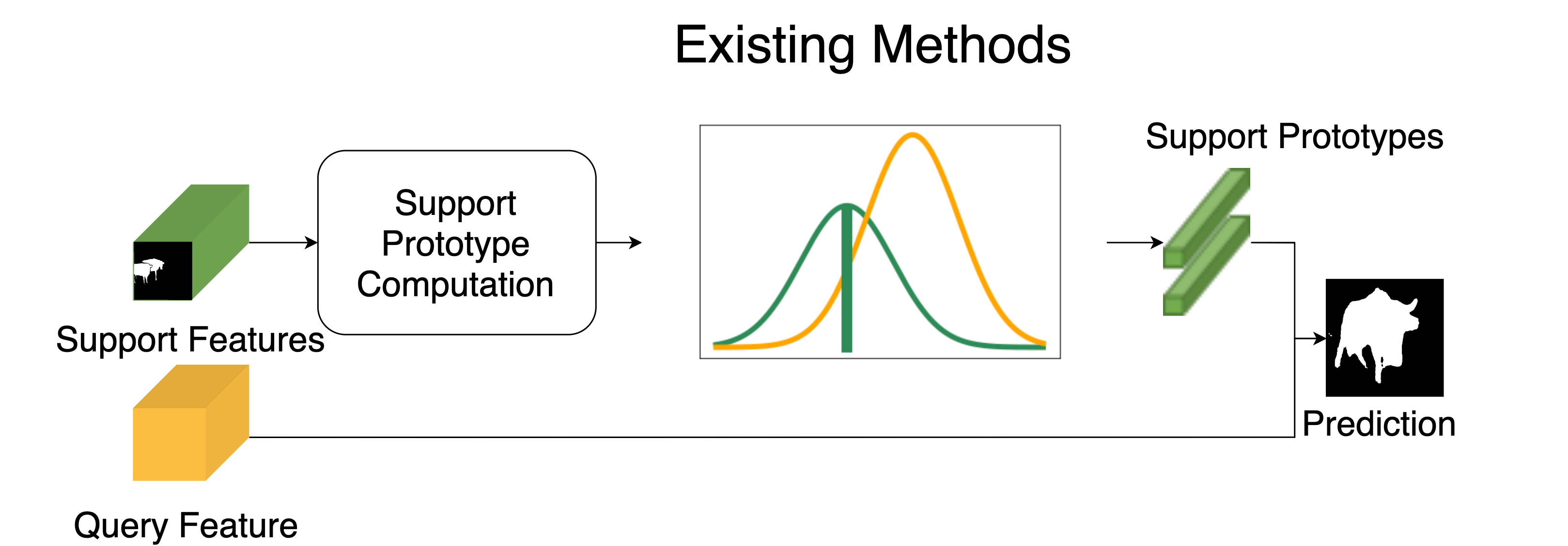}
\includegraphics[width=3.3in, height=1.15in]{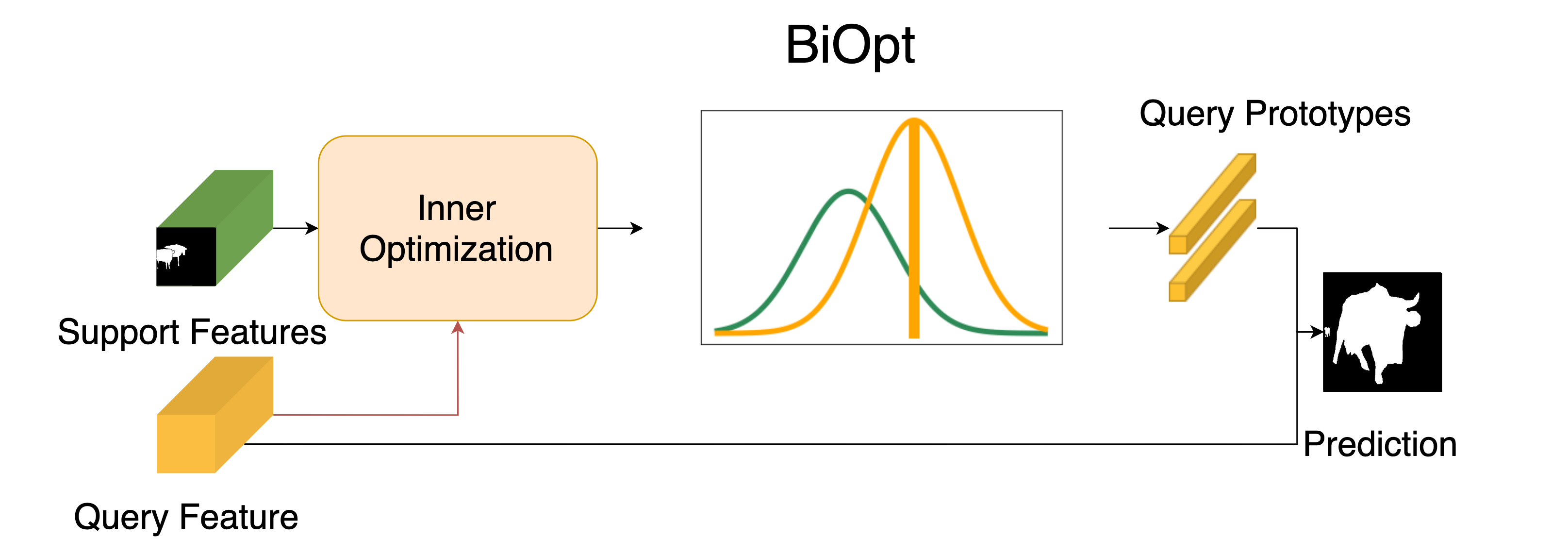}
\caption{Existing methods (upper) focus on extracting prototypes on the support images. BiOpt (lower) targets to estimate query prototypes which leverages semantic information on the unlabeled query image by inner optimization.}
\label{figure:abstract}
\end{figure}

\begin{figure}
\centering
\includegraphics[width=2.5in, height=1.5in]{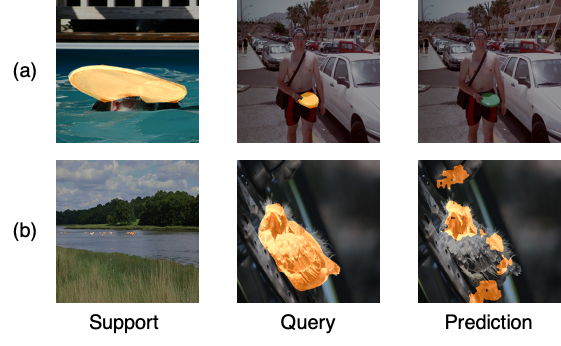}
\caption{Appearance and scale variation of support and query images in 1-shot episodes. Baseline model, where prototypes are extracted from the support images, fails to segment foreground objects on the query image. Best viewed in color with zoom in.}
\label{figure:case}
\end{figure}

Few-shot segmentation methods with prototype learning can be split into two branches. One is to concatenate support prototypes with query feature maps and output predicted masks through a segmentation head such as ASPP \cite{zhang2019canet, zhang2019pyramid, PMMs2020}. The other treats few-shot segmentation as the problem of pixel-wise classification. Feature vectors at each spatial location are assigned class labels of the nearest support prototype \cite{Wang_2019_ICCV, liu2020part}. 
These approaches commonly learn prototypes from the support images, Fig. \ref{figure:abstract} (upper). The extracted support prototypes are deemed as the discriminative representation for foreground and background classes. Then the query image is segmented by pixel-wise comparison between feature vectors and support prototypes. 
It is problematic when object appearances are significantly different among support and query images, especially in 1-shot scenarios displayed in Fig. \ref{figure:case}. Large foreground regions on the query image can not be segmented since the support prototypes lack semantic information on the query image. It is crucial to leverage knowledge on the query image into prototype estimation. However, due to the unavailability of query mask, existing methods fail to directly learn semantic knowledge on the unlabeled query image. 


\begin{figure}
\centering
\includegraphics[width=3in, height=1.6in]{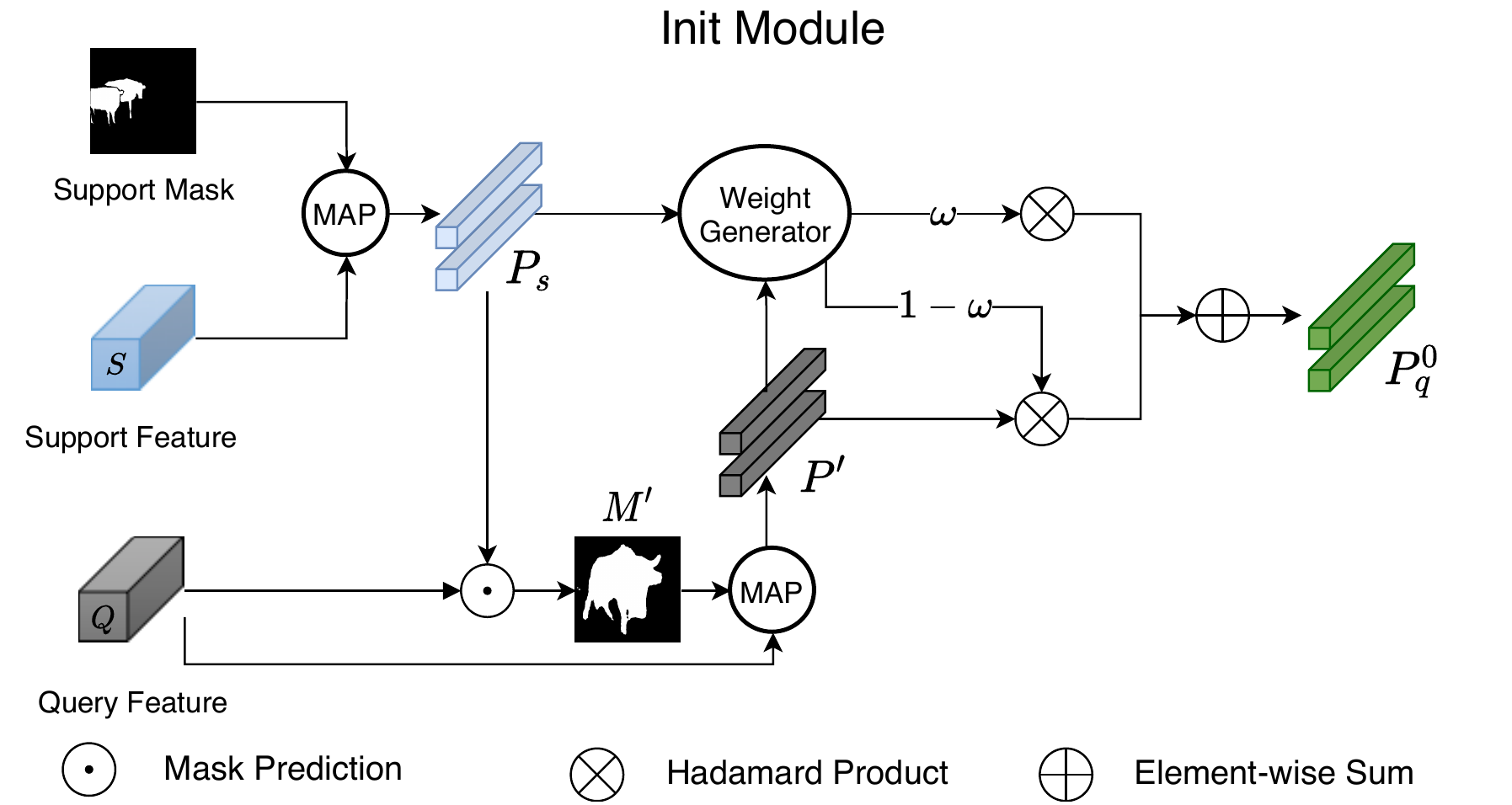}
\caption{The init module fully exploits knowledge from both support and query features, providing reasonably initialized query prototypes $P^0_q$ into the inner loop. MAP denotes the masked average pooling operation.}
\label{figure:init}
\end{figure}

In this paper, we propose the Bi-level Optimization (BiOpt) method to inductively estimate query prototypes.
Which is to say that, given a task $T=<(I_s, \mathcal{M}_s), I_q>$, we propose to extract query prototypes $P_q$ on the unlabeled query image $I_q$ in inductive setting. No additional unlabeled images are used in prototype estimation. 
Fig. \ref{figure:abstract} (lower) gives an overview of BiOpt, clearly showing the difference between recent approaches and BiOpt. 
They focus on directly extracting discriminative support prototypes $P_s$ on the support images such as prototype mixture \cite{PMMs2020} and part-aware prototype generation \cite{liu2020part}.
In contrast, BiOpt can inductively estimate query prototypes that leverage semantic information of target regions on the query image. Support images are used to guide prototype initialization during optimization, Fig. \ref{figure:init}.
Learning target is decomposed into two levels: inner objective to estimate query prototypes $P_q$ on each task and outer objective to learn a discriminative embedding space $\phi$ across tasks.

\begin{figure*}
\centering
\includegraphics[width=5.3in, height=3.45in]{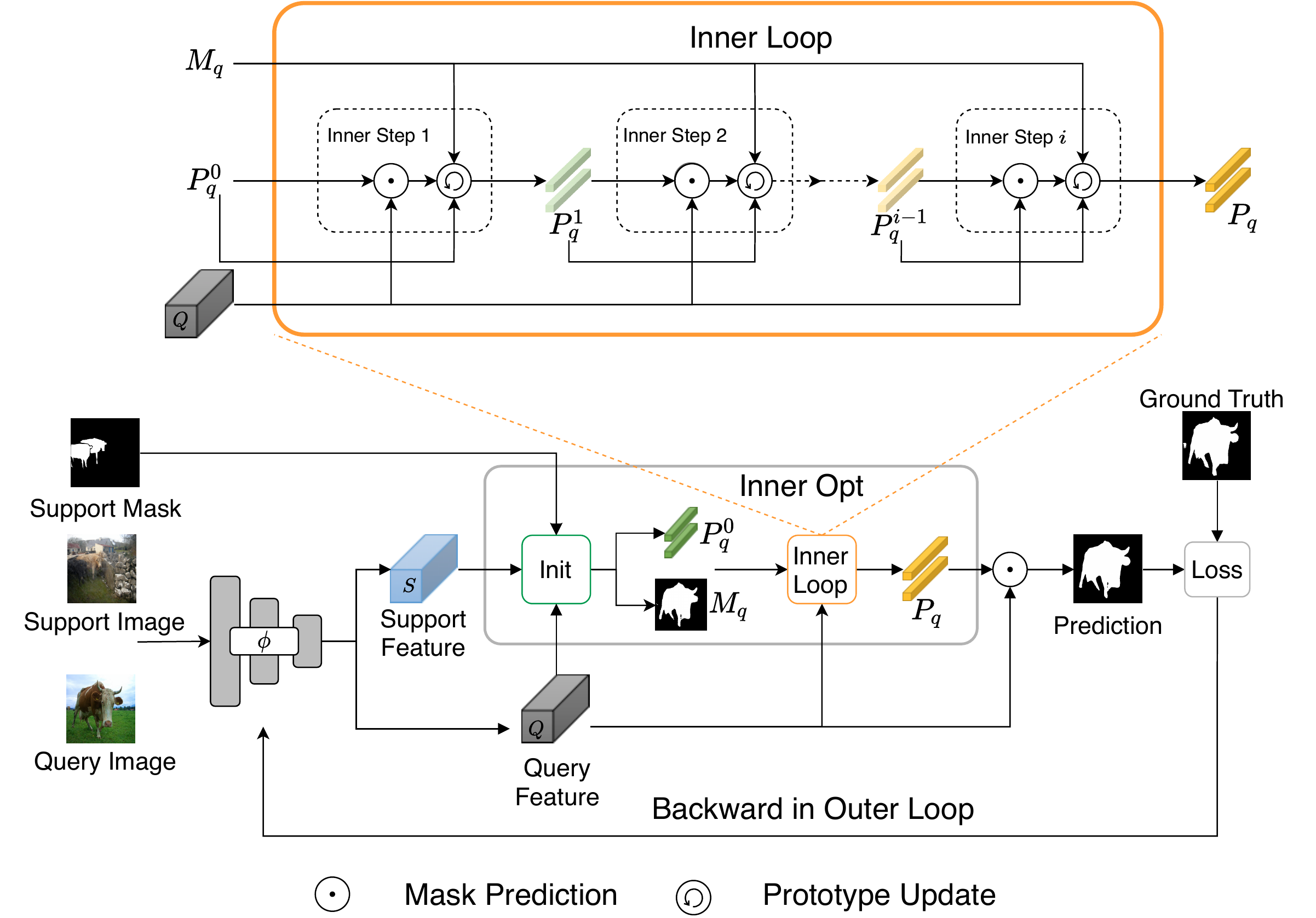}
\caption{End-to-end training framework of Bi-level Optimization (BiOpt). BiOpt is capable of leveraging semantic knowledge from the query image into prototype learning. In the inner loop, reasonably initialized query prototypes and target mask are first given by the init module. Then the query prototypes $P_q$ are learnt by iterative optimization. In the outer loop, $\phi$ is optimized across different tasks to learn discriminative embeddings.}
\label{figure:framework}
\end{figure*}

Query prototypes are supposed to have satisfying segmentation performance which can be directly measured by the segmentation loss. Based on this consideration, we expect to learn query prototypes by minimizing cross-entropy loss on the query image. Difficulty in inner optimization is the unavailability of query mask, aka the inner target. Furthermore, how to initialize query prototypes is another key issue in the inner loop. An init module is accordingly designed to initialize query prototypes and set an inner target for inner optimization reasonably. Then we iteratively update query prototypes by gradient descent. Optimized after inner procedure, our model can leverage knowledge from the unlabeled query image in prototype estimation. Outer optimization is further applied to learn a discriminative embedding space across different tasks. We follow the common practice that minimizes cross-entropy loss between predicted mask and ground-truth on the query images. Final prediction is obtained by annotating each vector on the query feature map with the class label of the nearest query prototype estimated in the inner loop. BiOpt enables to train models in an end-to-end manner and the train procedure is built in nested loops, Fig. \ref{figure:framework}.
The proposed inner optimization mechanism is feasible which can be directly applied at test.

We validate our method on PASCAL-$5^i$ and COCO-$20^i$ (including split-A and split-B). To our best knowledge, we are the first to give comprehensive results on all splits with three backbones. More than that, our BiOpt algorithm consistently outperforms the state-of-the-arts by large margins. Main contributions of this work are summarized as follows:

\begin{itemize}
  \item [1)] We propose a novel method BiOpt for inductive few-shot segmentation. The model can leverage semantic knowledge from the query image, which differs from most existing methods that focus on extracting prototypes only from the support images.
  
  \item [2)] We design a nested optimization algorithm to estimate the query prototypes on each task. In the inner loop, both support and query features are first exploited for good prototype initialization, and prototypes are optimized iteratively on the unlabeled query image. In the outer loop, we use the query prototypes obtained through inner optimization for prediction to optimize feature extractor. 
  
  \item [3)] Extensive experiments on PASCAL-$5^i$ and COCO-$20^i$ demonstrate the superiority of our proposed method. BiOpt achieves state-of-the-art results on 1-shot COCO-$20^i$ and 5-shot PASCAL-$5^i$ under all settings. 

\end{itemize}

\section{Related Work}

\textbf{Semantic Segmentation}
aims to associate each pixel in the image to a category label based on semantic understanding. 
FCN \cite{long2015fully} first proposed the fully convolutional architecture which replaces the final fully connected layer with the convolutional layer.
To obtain more accurate prediction, they also propose to use the deconvolution layer to increase the size of predicted masks.
U-Net \cite{ronneberger2015u} used a symmetric encoder-decoder architecture to better extract low level features.
In DeepLab-v1 \cite{chen14semantic}, the dilated convolution was proposed to reduce the information loss caused by the pooling layer. 
In order to better segment objects of different scales, DeepLab-v2 \cite{chen2017deeplab} proposed the ASPP module to fuse the features extracted from different scales. 
These semantic segmentation methods rely on much pixel-wisely annotated training images that are usually expensive and hard to obtain.

\textbf{Few-Shot Learning}
targets to recognize novel classes from limited training images \cite{vinyals2016matching,snell2017prototypical,sung2018learning,lee2019meta}.
ProtoNet \cite{snell2017prototypical} suggested that each class can be represented by a prototype in the feature space.
The classification problem can be performed by assigning label to the nearest prototype.
MAML \cite{finn2017model} and Reptile \cite{nichol2018first} proposed to learn a good model initialization which enables fast adaptation on novel tasks.
Optimization procedure is similarly designed in these methods. Optimal solution of each sub task is found in inner optimization and the model is trained through outer loop for generalization on different tasks.
This optimization procedure is also adopted by MetaOptNet \cite{lee2019meta}. In the inner loop, it finds optimal classifier, and in outer loop, feature extractor is optimized given the classification weights.
Inspired by these meta learning methods, we use bi-level optimization to solve few-shot segmentation problem.

\textbf{Few-Shot Segmentation}
targets to perform semantic segmentation in few-shot scenarios. 
This problem is recently proposed by OSLSM \cite{shaban2017one}. It provided a baseline in the few-shot segmentation community by using FCN for dense pixel-level prediction on a test image.
PL \cite{dong2018few} suggested to solve few-shot segmentation with prototype learning based on ProtoNet \cite{snell2017prototypical}. Recent prototype based methods \cite{PMMs2020, liu2020part} point out the limitation of single prototype in segmenting query images.
focus on learning prototypes by correlating diverse object parts. RPMMs \cite{PMMs2020} mixed prototypes by Expectation-Maximization algorithm to correlate diverse object parts on the support images. PPNet \cite{liu2020part} decomposed the single class representation into a set of part-aware prototypes to capture fine-grained parts on support features. PANet \cite{Wang_2019_ICCV} bi-directionally estimated prototypes to learn a consistent embedding space.
It can be seen that existing methods focus on exploiting knowledge from the support images.
We find that support prototypes show unsatisfying segmentation performance when handling large appearance and scale variation between support and query objects.
To deal with it, we propose to estimate query prototypes by leveraging semantic information on the unlabeled query image.

\section{Problem Setup}
In few-shot segmentation, we aim to learn a model that can perform segmentation on novel objects from limited labeled images per class.
The problem of few-shot segmentation is formed in meta tasks, also called episodes. 
Training tasks $\mathcal{T}^{tr}$ are constructed from the base classes $\mathcal{C}_{base}$ and test tasks $\mathcal{T}^{te}$ are sampled from the novel classes $\mathcal{C}_{novel}$. Two class sets are disjoint which is to say that, $\mathcal{C}_{novel}$ remains unseen before the test.
Each task is formally composed of a support set $\mathcal{S}$ and a query set $\mathcal{Q}$.
In the $N$-way $K$-shot setting, the support set $\mathcal{S}=\{(I_s, \mathcal{M}_s)\}$ contains $K$ support images $I_s\in \mathbbm{R}^{H\times W\times 3}$ with ground-truth masks $\mathcal{M}_s\in \mathbbm{R}^{H\times W\times 1}$ for each of $N$ classes. The query set $\mathcal{Q}=\{(I_q, \mathcal{M}_q)\}$ has $N_q$ query images $I_q$ from these $N$ classes, with ground-truth masks $\mathcal{M}_q$ available in training but unknown in test.
The segmentation model is trained episodically on $\mathcal{T}^{tr}$, learning to perform segmentation in few-shot scenarios. After episodic training, the model is test on $\mathcal{T}^{te}$ for evaluation.

\section{Bi-Level Optimization}
In few-shot segmentation, the general objective of prototype based methods is to learn an embedding space $\phi$ where emperical segmentation loss is minimum given a base learner $\mathcal{A}$. Learning objective can be formulated as:
\begin{equation}
\min_{\phi} \mathbb{E}_{\mathcal{T}^{tr}} [\mathcal{L}_{seg}( \mathcal{Q}; P, \phi) ],  {\rm {where}}  \;  P = \mathcal{A}(\mathcal{S}; \phi)
\end{equation}
The base learner $\mathcal{A}$ extracts prototypes $P$ on the support images with ground-truth masks by masked average pooling \cite{zhang2019canet}, prototype mixture \cite{PMMs2020} or part-aware prototype generation \cite{liu2020part}. Main problem solved in these approaches is how to extract representative prototypes on the \textit{support} images $I_s$. It will lead to unsatisfying segmentation performance when the appearance of objects on $I_s$ and $I_q$ varies significantly, Fig. \ref{figure:case}. Unlike these methods, we challenge to estimate query prototypes on the unlabeled query image into prototype computation, achieving satisfying results in the above mentioned situation.

\subsection{Our Objective}
Our objective is to extract query prototypes in inductive setting, without importing additional unlabeled images as in \cite{liu2020part}. 
Since the query masks are unknown in test, the model should be capable of leveraging knowledge on the unlabeled query image.
The nested training process is designed where inner loops are accordingly set to mimic the test scenarios at the training stage. 
The objective is decomposed into two levels and achieved in corresponding loops.
1) \textbf{The inner objective} is to learn query prototypes \textbf{$P_q$} from the unlabeled query image on each task. 
2) \textbf{The outer objective} is to learn a discriminative embedding space \textbf{$\phi$} across tasks.

\subsubsection{Inner Objective}
In this paper, we propose to build a base learner which can directly estimate prototypes from the unlabeled query image:
\begin{equation}
\begin{aligned}
P = \mathcal{A}(\mathcal{Q}; \mathcal{S}, \phi) 
\end{aligned}
\label{eq:newprototype}
\end{equation}
where $P\in \mathbbm{R}^{C\times (N+1)}$ includes background class.
Existing approaches aim to obtain prototypes that have the minimum distance to feature vectors in the region or subregion of target classes, on the support images. Differently, we expect the prototypes to have the minimum segmentation loss on the query image which is an explicit indicator of segmentation result. The objective is formulated as:
\begin{equation}
\begin{aligned}
\min_{P_q} \sum_{(x,y)} \mathcal{L}_{CE}(softmax (\, Cos(Q^{(x,y)}, P_q), {\mathcal{M}}_q^{(x,y)}) )
\end{aligned}
\label{eq:innerobjective}
\end{equation}
where $(x,y)$ are spatial locations on feature maps $Q = \phi (I_q) \in \mathbbm{R}^{H\times W\times C}$.
$Cos(\cdot)$ is cosine similarity operator and $\mathcal{L}_{CE}$ is cross-entropy loss.
Note the ground-truth $\mathcal{M}_q$ is unavailable in the inner loop. It is consistent with test scenarios.
Therefore, we use $\mathit{M}_q \in \mathbbm{R}^{H\times W\times 1}$ as substituted ground-truth in inner optimization. 
The way to build target mask $\mathit{M}_q$ is introduced in the next subsection.

\subsubsection{Outer Objective}
Few-shot segmentation model is expected to generalize well to novel classes. It is usually solved by minimizing empirical error on base classes. Our model is optimized across tasks $\mathcal{T}^{tr}$ over base classes with the outer objective:
\begin{equation}
\begin{aligned}
\min_{\phi} \ \sum_{\mathcal{T}^{tr}} \mathcal{L}_{CE}(softmax(\, Cos(\phi (I_q), P_q)), \mathcal{M}_q)
\end{aligned}
\label{eq:outerobjective}
\end{equation}
where $P_q$ is given after the inner loop and $\mathcal{M}_q$ is the ground-truth mask.
Through outer optimization, our model can learn a good embedding space where features are discriminative for different classes.

\subsection{BiOpt Algorithm}
\label{sec:algo}
Framework of our BiOpt is shown in Fig. \ref{figure:framework} and the algorithm is described in detail in this subsection. 
BiOpt trains a segmentation model end-to-end through nested loops.

\textbf{Inner Optimization}
Given embedding function $\phi$, usually a deep neural network, there are two variables $P_q$ and $\mathit{M}_q$ to be determined in the inner objective. Query prototype $P_q$ is the variable to be optimized in the inner loop and $\mathit{M}_q$ is the substituted target mask in loss computation. How to initialize $P_q$ before inner optimization and how to set the target are two problems we solve in the following. 
We propose an init module as displayed in Fig. \ref{figure:framework} for initializing query prototypes. It fully exploits semantic knowledge from annotated support images and unlabeled query images to find a good initialization. 
The support prototypes $P_s$ are first computed by masked average pooling, denoted by MAP operator in figure. Temporary query mask $M'$ can be given by pixel-wise comparison with $P_s$ and temporary query prototypes $P'$ can be similarly computed. $M'$ and $P'$ are roughly computed but contains discriminative information on the query image to some extent. To leverage these information without importing much noises, we feed $P'$ and $P_s$ into a weight generator and initialize query prototypes as:
\begin{equation}
P^0_q = \omega \cdot P_s + (1-\omega) \cdot P'
\label{eq:initialization}
\end{equation}
where $w\in \mathbb{R}^{C\times (N+1})$ is generated from weight generator which consists of a concatenation operator, a fully-connected layer and a sigmoid function. 
Consequently, the target mask can be set as:
\begin{equation}
\mathit{M}_q^{(x,y)}  = \arg \max_{c} \; \frac{exp(-\alpha \cdot Cos(Q^{(x,y)}, P_{q,c}^0))}{\sum_{c\in\mathcal{C}^T} exp(-\alpha \cdot Cos(Q^{(x,y)}, P_{q,c}^0))}
\label{eq:targetmask}
\end{equation}
where $\alpha$ is a scalar parameter fixed at 20 at training. $\alpha$ is omitted in other equations of mask prediction for brevity. $\mathcal{C}^T$ is the semantic class space on task $T$.
We can set better initialization and target mask in this way and verify it in experiments.

Given initialized query prototypes $P^0_q$ and target mask $\mathit{M}_q$, we iteratively update prototypes by gradient descent in the inner loop.
\begin{equation}
P^i_q = P^{i-1}_q -  \nabla \mathcal{L}_{CE}( softmax (\, Cos(Q, P^{i-1}_q)), \mathit{M}_q)
\label{eq:innerupdate}
\end{equation}
Optimized query prototypes $P_q$ approximate to the expected prototypes by ensuring small segmentation loss on the query image.

\textbf{Outer optimization}
Final prediction $\hat{\mathcal{M}}_q$ is similarly computed as in Eq. \ref{eq:targetmask} where prototypes are substituted by optimized $P_q$.
To strengthen regularization of the training process, we impose constraints on two intermediate results $M'$ and $\mathit{M}_q$. The model is overall trained by minimizing segmentation loss as:
\begin{equation}
\mathcal{L}_{seg} = \mathcal{L}_{CE}(M', \mathcal{M}_q) + \mathcal{L}_{CE}(\mathit{M}_q, \mathcal{M}_q) + \mathcal{L}_{CE}(\hat{\mathcal{M}}_q, \mathcal{M}_q) 
\end{equation}

Our BiOpt is feasible at both training and test phases. 
At the training stage, the model is trained end-to-end. It can learn query prototypes in the inner loop where tasks are set to mimic test scenarios. 
At the test stage, we apply inner optimization on each task and infer by the optimized query prototypes.

The proposed BiOpt algorithm for few-shot segmentation is summarized in Algorithm \ref{alg}. 
\begin{algorithm}
\caption{Bi-Level Optimization (BiOpt)}
\label{alg}
\begin{algorithmic}

\REQUIRE {Tasks $\mathcal{T}^{tr}=\{(I_s, I_q, \mathcal{M}_s, \mathcal{M}_q)\}$ on $\mathcal{C}_{base}$}
\STATE {Initialize feature extractor $\phi$} 
\FOR{j in Outer Loops}
\STATE {Get feature maps $S$, $Q$ = $\phi(I_s)$, $\phi(I_q)$}

\STATE {Compute $P_s $, $M'$, $P'$, and generate weight $\omega$}

\STATE {Initialize query prototypes $P^0_{q}$ by Eq. \ref{eq:initialization}}

\STATE {Set target mask $\textit{M}_q$ by Eq. \ref{eq:targetmask}}

	\FOR{i in Inner Loops}
		\STATE{Update $P_q^i$ by Eq. \ref{eq:innerupdate}}
			
	\ENDFOR
	
	Compute $\hat{\mathcal{M}}_q = softmax(\, Cos(Q, P_q)) $
	
\STATE {Update $\phi^j =  \phi^{j-1} - \nabla \mathcal{L}_{seg}$  } 

\ENDFOR
\end{algorithmic}
\end{algorithm}

\section{Experiments}
\subsection{Setup}
\textbf{Dataset}
Two benchmarks PASCAL-$5^i$ \cite{shaban2017one} and COCO-$20^i$ \cite{nguyen2019feature} are utilized in this paper. PASCAL-$5^i$ is derived from PASCAL VOC 2012 \cite{everingham2010the} which is split into 4 disjoint folds with 5 categories per fold. We use the dataset provided by \cite{zhang2019canet} in experiments. 
COCO-$20^i$ is a more challenging dataset derived from MS COCO \cite{lin2014microsoft}. We denote the split proposed in \cite{nguyen2019feature} as split-A while the split in \cite{hu2019attention} as split-B. 80 classes are divided into 4 folds in both splits. Results are reported on all splits in this paper. All images are resized to 417$\times$417 in experiments.

\textbf{Network}
We conduct comprehensive experiments on three networks: VGG-16 \cite{simonyan2015very}, ResNet-50 and ResNet-101 \cite{he2016deep}. Following previous works \cite{shaban2017one,Wang_2019_ICCV}, networks are initialized by the pretrained weights on ILSVRC \cite{russakovsky2015imagenet}. In the last two blocks in ResNets, strides of the last two blocks are set to 1 and dilated convolutions with rates of 2 and 4 are used for larger reception fields. Layers after the 4-th block and the last ReLU layer are removed for prototype computation.

\textbf{Implementation Detail}
We train 10,000 epochs on PASCAL-$5^i$ and 20,000 epochs on COCO-$20^i$. Batch size is set to 8 in 1-shot experiments and 2 in 5-shot experiments.
In outer optimization, our model is trained by SGD with the learning rate of 7e-3, momentum of 0.9 and weight decay of 5e-4. The learning rate is reduced by 0.1 after 10,000 iterations on COCO derivatives. 1,000 episodes are randomly sampled for evaluation. $N_q = 1$ in all experiments. At both training and evaluation phases, the learning rate in inner optimization is 0.1 and the number of inner steps is set to 10. Random crop is used for data augmentation as in \cite{zhang2019canet}. 

\textbf{Evaluation}
In few-shot segmentation, the model is trained on three folds and evaluated on the rest fold.
Mean-IoU and binary-IoU are frequently used evaluation protocols. Mean-IoU measures the averaged IoU score on all foreground classes. Binary-IoU treats all foreground objects as one class and the background is viewed as one class.

\textbf{Baseline}
We train a baseline model for comparison to show the effectiveness of bi-level optimization. It extracts prototypes from the support images by masked average pooling and directly predicts on the query image.

\begin{table*}[]
\centering
\begin{tabular}{lccccccccccc}
\toprule
\multirow{2}{*}{\textbf{Methods}} & \multirow{2}{*}{\textbf{Backbone}} & \multicolumn{5}{c}{\textbf{1-shot}}   &  \multicolumn{5}{c}{\textbf{5-shot}}      \\
 &  & Fold 1      & Fold 2      & Fold 3      & Fold 4      & {\color[HTML]{D34F00}  \textbf{Mean} }       & Fold 1      & Fold 2      & Fold 3      & Fold 4      & {\color[HTML]{D34F00} \textbf{Mean}}         \\
\hline
OSLSM \cite{shaban2017one}            & \multirow{7}{*}{VGG-16}     & 33.60                & 55.30                & 40.90                & 33.50                & {\color[HTML]{D34F00} 40.80}                & 35.90                & 58.10                & 42.70                & 39.10                & {\color[HTML]{D34F00} 43.95}                 \\
co-FCN \cite{rakelly2018conditional}           &                             & 36.70                & 50.60                & 44.90                & 32.40                & {\color[HTML]{D34F00} 41.10}              & 37.50                & 50.00                & 44.10                & 33.90                & {\color[HTML]{D34F00} 41.40}                \\
SG-One \cite{zhang2020sg}           &                             & 40.20                & 58.40                & 48.40                & 38.40                & {\color[HTML]{D34F00} 46.30  }              & 41.90                & 58.60                & 48.60                & 39.40                & {\color[HTML]{D34F00} 47.10}               \\
PANet \cite{Wang_2019_ICCV}           &                             & 42.30                & 58.00                & 51.10                & 41.20                & {\color[HTML]{D34F00} 48.10 }               & 51.80                & 64.60                & 59.80                & 46.50                & {\color[HTML]{D34F00} 55.70}       \\
FWB \cite{nguyen2019feature} &	& 47.04                & 59.64                & 52.51                & 48.27                & {\color[HTML]{D34F00} 51.90 }               & 50.87                & 62.86                & 56.48                & 50.09                & {\color[HTML]{D34F00} 55.08}                  \\
RPMMs \cite{PMMs2020}            &                             & 47.14                & 65.82                & 50.57                & 48.54               &  {\color[HTML]{D34F00} 53.02 }                & 50.00                & 66.46                & 51.94                & 47.64                & {\color[HTML]{D34F00} 54.01}            \\
\textbf{BiOpt}        &                             & 50.01                & 63.41                & 56.73                & 43.60                &  {\color[HTML]{D34F00} \textbf{53.44} }           & 56.27                & 65.65                & 63.15                & 49.42                & {\color[HTML]{D34F00} \textbf{58.62} }             \\
\hline
CANet \cite{zhang2019canet}            & \multirow{5}{*}{ResNet-50}  & 52.50                & 65.90                & 51.30                & 51.90                & {\color[HTML]{D34F00} 55.40 }                & 55.50                & 67.80                & 51.90                & 53.20                & {\color[HTML]{D34F00} 57.10}               \\
PGNet \cite{zhang2019pyramid}            &                             & 56.00                & 66.90                & 50.60                & 50.40                & {\color[HTML]{D34F00} 56.00}                & 57.70                & 68.70                & 52.90                & 54.60                & {\color[HTML]{D34F00} 58.50}               \\
RPMMs \cite{PMMs2020}            &                             & 55.15                & 66.91                & 52.61                & 50.68                & {\color[HTML]{D34F00} 56.34}                & 56.28                & 67.34                & 54.52                & 51.00                & {\color[HTML]{D34F00} 57.30}                \\
PPNet \cite{liu2020part}            &                             & 48.58                & 60.58                & 55.71                & 46.47                & {\color[HTML]{D34F00} 52.84}                & 58.85                & 68.28                & 66.77                & 57.98                & {\color[HTML]{D34F00} 62.97}            \\
\textbf{BiOpt}        &                             & 56.21                & 64.63                & 57.95                & 48.61                & {\color[HTML]{D34F00} \textbf{56.85}  }          & 61.32                & 69.45                & 67.74                & 58.11                & {\color[HTML]{D34F00} \textbf{64.16} }                \\
\hline
FWB \cite{nguyen2019feature}              & \multirow{4}{*}{ResNet-101} & 51.30                & 64.49                & 56.71                & 52.24                & {\color[HTML]{D34F00} 56.19}                & 54.84                & 67.38                & 62.16                & 55.30                & {\color[HTML]{D34F00} 59.92}          \\
PPNet \cite{liu2020part}            &                             & 52.71                & 62.82                & 57.38                & 47.74                & {\color[HTML]{D34F00} 55.16}                & 60.25                & 70.00                & 69.41                & 60.72                & {\color[HTML]{D34F00} 65.10}                      \\
DAN \cite{wang2020few}              &                             & 54.70                & 68.60                & 57.80                & 51.60                & {\color[HTML]{D34F00} \textbf{58.20} }           & 57.90                & 69.00                & 60.10                & 54.90                & {\color[HTML]{D34F00} 60.50}                \\
\textbf{BiOpt}         &                             & 58.16                & 64.82                & 59.12                & 49.23                & {\color[HTML]{D34F00} 57.83}                & 62.72                & 73.27                & 68.20                & 57.00                & {\color[HTML]{D34F00} \textbf{65.30} }      \\
\bottomrule
\end{tabular}
\caption{Mean-IoU results on 1-way \textbf{PASCAL-$5^i$}. The best result over four folds is marked in bold.}
\label{tb:pascal}
\end{table*}

\subsection{Comparison with the State-of-the-arts}

To give a fair comparison with the state-of-the-arts, we conduct comprehensive experiments on PASCAL-$5^i$ and COCO-$20^i$ of two splits. In addition, we report detailed results with three backbones. Mean-IoU results are displayed in Table \ref{tb:pascal} and Table \ref{tb:coco}.

\subsubsection{PASCAL-$5^i$}
We provide comprehensive mean-IoU results in Table \ref{tb:pascal} to fairly compare with the state-of-the-arts using the same backbones. 
Model performance is usually measured by the mean results over four folds which are denoted in orange color.
It can be seen from Table \ref{tb:pascal} that our method achieves the best performance in most settings, except for 1-shot with ResNet-101.
With VGG-16 as the embedding network, BiOpt outperforms state-of-the-arts by a margin of 0.42\% in 1-shot and 2.92\% in 5-shot. 
PANet \cite{Wang_2019_ICCV} proposed the alignment loss to exploit the query image. We improve it by 5.34\% which demonstrates that BiOpt can estimate better query prototype by proposed inner optimization.
Recent literature mainly reports results on the ResNet-50 network, which has better performance than VGG-16.
Compared with existing methods with ResNet-50 as backbone, BiOpt improves the segmentation performance by 0.51\% and 1.19\% on 1-shot and 5-shot tasks respectively.
RPMMs \cite{PMMs2020} and PPNet \cite{liu2020part} are recently proposed to learn support prototypes by mixing semantic information from spatial parts on the support images. In comparison with these methods, performance of BiOpt is superior which indicates the effectiveness of estimating query prototypes.
Table \ref{tb:biou} gives binary-IoU results with ResNet-50. We achieve 70.74\% (1-shot) and 75.66\% (5-shot) on PASCAL-$5^i$.

\subsubsection{COCO-$20^i$}
Table \ref{tb:coco} compares the results on COCO-$20^i$ dataset with existing methods.
On COCO-$20^i$-A, we achieve the state-of-the-art performance in 1-shot setting 
BiOpt brings improvements by 2.18\% (VGG-16) and 5.9\% (ResNet-50) compared with PANet. 
On COCO-$20^i$-B, our method is superior to existing methods on all backbones.
PPNet \cite{liu2020part} and RPMMs \cite{PMMs2020} provide competitive results by exploiting a set of support prototypes.
However, BiOpt shows better performance especially in 1-shot, which achieves 26.96\%, 33.15\%, 34.57\% respectively.
BiOpt is the first work to provide binary-IoU results on COCO-$20^i$ of two splits, as shown in Table \ref{tb:biou}.
It serves as a baseline for future research.

\begin{table*}[]
\centering
\begin{tabular}{lllllllllllc}
\toprule
 \multirow{2}{*}{\textbf{Methods}} & \multirow{2}{*}{\textbf{Backbone}} & \multicolumn{5}{c}{\textbf{1-shot}}   &  \multicolumn{5}{c}{\textbf{5-shot}}      \\
 & &  Fold 1      & Fold 2      & Fold 3      & Fold 4      & {\color[HTML]{D34F00}  \textbf{Mean} }       & Fold 1      & Fold 2      & Fold 3      & Fold 4      & {\color[HTML]{D34F00} \textbf{Mean}}         \\
\hline
 \multicolumn{12}{c}{split-A} \\
 \hline
PANet \cite{Wang_2019_ICCV}                                       & VGG-16            & 28.70           & 21.20           & 19.10           & 14.80           & {\color[HTML]{D34F00} 20.90}         & 39.43           & 28.30           & 28.20           & 22.70           & {\color[HTML]{D34F00} \textbf{29.70}}         \\
PANet \cite{Wang_2019_ICCV}*                                     & ResNet-50         & 31.50           & 22.58           & 21.50           & 16.20           & {\color[HTML]{D34F00} 22.95}         & 45.85           & 29.15           & 30.59           & 29.59           & {\color[HTML]{D34F00} 33.80}         \\
PPNet \cite{liu2020part}                                       & ResNet-50         & 36.48           & 26.53           & 25.99           & 19.65           & {\color[HTML]{D34F00} 27.16}         & 48.88           & 31.36           & 36.02           & 30.64           & {\color[HTML]{D34F00} \textbf{36.73}}         \\ 
\hline
 \multicolumn{1}{c}{}                       & VGG-16            & 32.95           & 24.29           &       18.58          & 16.51           & {\color[HTML]{D34F00} \textbf{23.08} }              & 38.08           & 27.52           & 27.30           & 21.83           & {\color[HTML]{D34F00} 28.68}         \\
 \multicolumn{1}{c}{}                       & ResNet-50         & 38.63           & 29.61           & 23.52           & 23.62           & {\color[HTML]{D34F00} \textbf{28.85}}         & 47.44           & 34.42           & 31.89           & 27.90           & {\color[HTML]{D34F00} 35.41}         \\
\multirow{-3}{*}{\textbf{BiOpt}}           & ResNet-101        & 41.55           & 29.83           & 25.30           & 24.74           & {\color[HTML]{D34F00} \textbf{30.36}}         & 49.22           & 33.81           & 31.47           & 33.25           & {\color[HTML]{D34F00} \textbf{36.94}}         \\
\hline
\multicolumn{12}{c}{split-B} \\ 
\hline
PPNet \cite{liu2020part}                                       & ResNet-50         & 28.09           & 30.84           & 29.49           & 27.70           & {\color[HTML]{D34F00} 29.03}         & 38.97           & 40.81           & 37.07           & 37.28           & {\color[HTML]{D34F00} 38.53}         \\
RPMMs \cite{PMMs2020}                                      & ResNet-50         & 29.53           & 36.82           & 28.94           & 27.02           & {\color[HTML]{D34F00} 30.58}         & 33.82           & 41.96           & 32.99           & 33.33           & {\color[HTML]{D34F00} 35.52}         \\
FWB \cite{nguyen2019feature}                                       & VGG-16        &  18.35         & 16.72       &   19.59       &  25.43       & {\color[HTML]{D34F00} 20.02 }         &  20.94          &  19.24          &    21.94        &   28.39         & {\color[HTML]{D34F00} 22.63}         \\  
FWB \cite{nguyen2019feature}                                      & ResNet-101        & 16.98           & 17.98           & 20.96           & 28.85           & {\color[HTML]{D34F00} 21.19}         & 19.13           & 21.46           & 23.93           & 30.08           & {\color[HTML]{D34F00} 23.05}         \\  
\hline
                                           & VGG-16            & 24.61           & 30.84           & 25.96           & 26.41           & {\color[HTML]{D34F00} \textbf{26.96}}         & 29.29           & 34.90           & 31.87           & 30.14           & {\color[HTML]{D34F00} \textbf{31.55}}         \\
                                            & ResNet-50         & 32.23           & 36.05           & 32.00           & 32.33           & {\color[HTML]{D34F00} \textbf{33.15}}         & 37.96           & 44.43           & 36.24           & 37.12           & {\color[HTML]{D34F00} \textbf{38.94}}         \\
 \multirow{-3}{*}{\textbf{BiOpt}}                     & ResNet-101        & 34.28           & 38.50           & 34.75           & 30.76           & {\color[HTML]{D34F00} \textbf{34.57}}         & 38.49           & 46.18           & 35.81           & 38.98           & {\color[HTML]{D34F00} \textbf{39.87}}        \\
\bottomrule
\end{tabular}
\caption{Mean-IoU results on 1-way \textbf{COCO-$20^i$}. * Results are reported in \cite{liu2020part}. The best result with different backbone is marked in bold.}
\label{tb:coco}
\end{table*}

\begin{table}[]
\centering
\begin{tabular}{lccc}
\hline
Methods & Dataset & 1-shot  & 5-shot  \\
\hline
PANet \cite{Wang_2019_ICCV}  &  \multirow{5}{*}{PASCAL-$5^i$}    &  66.50   &  70.70    \\
CANet \cite{zhang2019canet}      &          &     66.20             &  69.60       \\
PGNet \cite{zhang2019pyramid}   &           &     69.90             &   70.50      \\
PPNet \cite{liu2020part}           &      &    69.19              &  75.76       \\
BiOpt         &   &     70.74             &     75.66    \\
\hline
BiOpt     & COCO-$20^i$-A            &       64.11           &    66.40     \\
BiOpt     & COCO-$20^i$-B        &        66.52          &  69.01      \\
\hline
\end{tabular}
\caption{Binary-IoU on PASCAL-$5^i$ and COCO-$20^i$. Compared with methods on ResNet-50.}
\label{tb:biou}
\end{table}

\subsection{Ablation Study}
In this subsection, we perform ablation study to inspect the effect of different components in our method. All experiments are implemented on ResNet-50.

\textbf{Inner Optimization}
We conduct experiments on 1-shot PASCAL-$5^i$ to evaluate the effectiveness of our bi-level optimization strategy in Table \ref{tb:innerablation}. Baseline results are shown in the first row where mean-IoU on four folds is 51.70\%. 
Bottom two lines show results of the baseline model equipped with our proposed inner optimization mechanism.
The performance is consistently improved by auxiliary optimization during the inner loop. Especially initialized through our init module, mean-IoU result is increased by a large margin of 5.15\% over four folds. It indicates that our model can estimate effective query prototype through inner optimization.

\begin{table}[]
\begin{tabular}{llccccc}
\toprule
Baseline & Fold 1 & Fold 2 & Fold 3 & Fold 4 & Mean \\
\hline
       			& 49.88           & 60.81           & 50.81           & 45.66           & 51.70         \\
+ InOpt$\dag$             & 53.69           & 62.43           & 54.74           & 48.17           & 54.76       \\
+ InOpt                 & 56.21           & 64.63           & 57.95           & 48.61           & 56.85       \\
\bottomrule
\end{tabular}
\caption{Ablation study on 1-shot PASCAL-$5^i$. InOpt$\dag$ refers to set $P_s$ as the initialized prototype and $M'$ as the target mask in the inner loop. InOpt is the inner optimization strategy described in the main paper.}
\label{tb:innerablation}
\end{table}

\begin{figure}
\centering
\includegraphics[width=2.9in, height=1.9in]{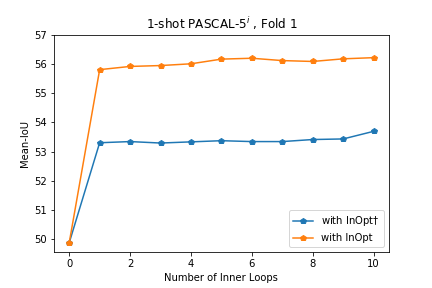}
\caption{Results with different numbers of inner steps.}
\label{tb:innerstep}
\end{figure}

\textbf{Inner Step}
Fig. \ref{tb:innerstep} shows mean-IoU results with different inner steps. Result with 0 inner loop is actually the result of the baseline model, 49.88\%. With step ranging from 1 to 10, performance of InOpt$\dag$ is raised from 53.29\% to 53.69\%, in the blue line. Similarly in the orange line, result  55.80\% to 56.21\%
We observe that our model shows stable performance towards different steps, free from fine-tuning hyper-parameters in experiments. Furthermore, the proposed inner optimization brings significant improvements over the baseline model.

\textbf{Multi-scale Test}
Following common practice in few-shot segmentation \cite{zhang2019pyramid, zhang2019pyramid}, the query image is re-scaled by [0.7, 1, 1.3] in the test. Multi-scale results on PASCAL-$5^i$ and COCO-$20^i$-A are provided in Table \ref{tb:multiscale}. We report mean scores over four folds on each dataset. The improvement caused by multi-scale evaluation is up to 0.84\% on 5-shot PASCAL-$5^i$. On 1-shot COCO-$20^i$-A, the result is merely raised by 0.09\% in multi-scale test.
\begin{table}[]
\centering
\begin{tabular}{lcccc}
\hline
\multicolumn{1}{c}{} & \multicolumn{2}{c}{PASCAL-$5^i$} & \multicolumn{2}{c}{COCO-$20^i$-A} \\
\multicolumn{1}{c}{} & 1-shot       & 5-shot      & 1-shot       & 5-shot      \\
\hline
w/o                  & 56.04        & 63.17       & 28.76        & 34.61       \\
w/                   & 56.85        & 64.16       & 28.85        & 35.41      \\
\hline
\end{tabular}
\caption{Multi-scale test on PASCAL-$5^i$ and COCO-$20^i$-A.}
\label{tb:multiscale}
\end{table}

\begin{figure*}
\centering
\includegraphics[width=6.6in, height=4.95in]{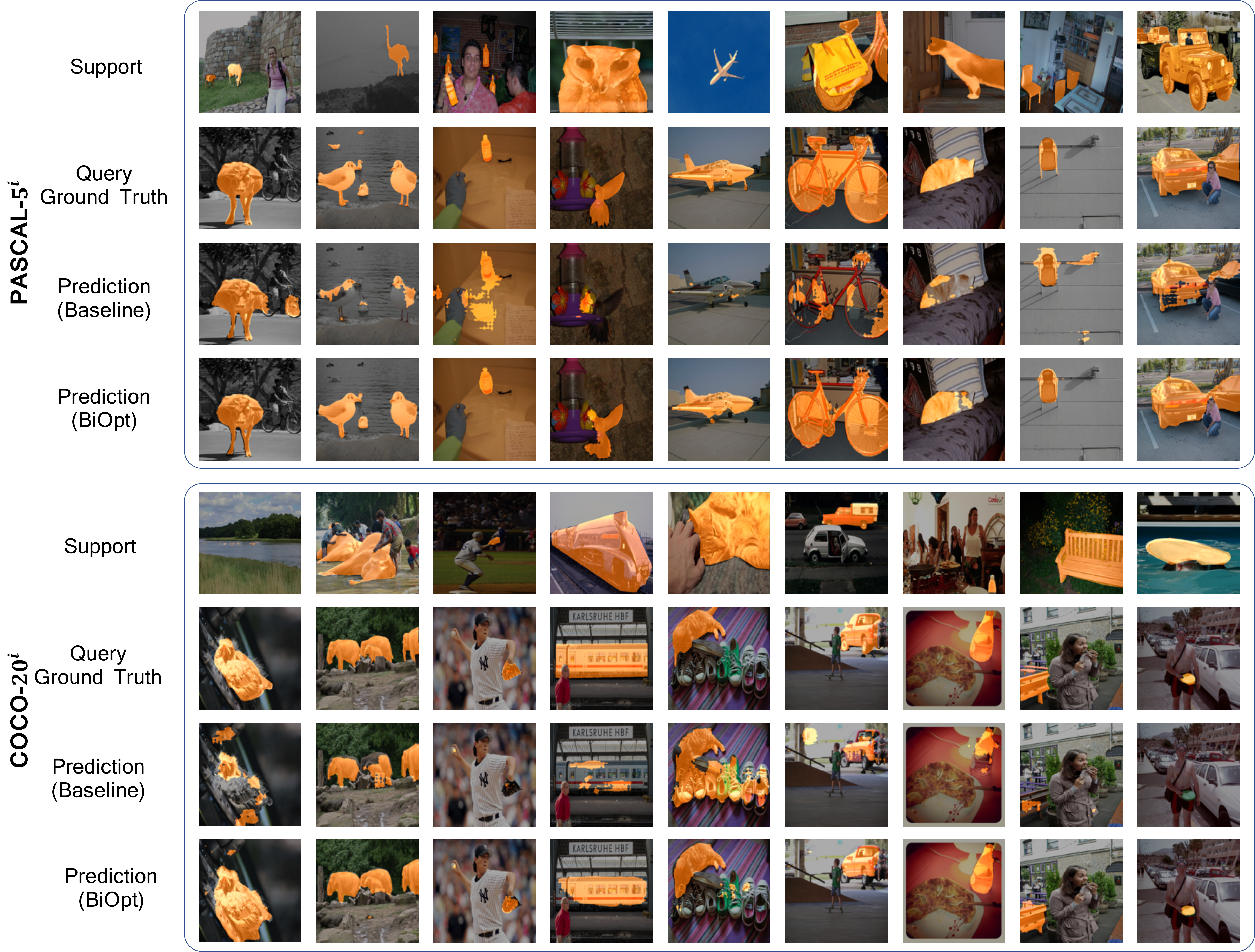}
\caption{Qualitative visualization of 1-shot episodes on PASCAL-$5^i$ and COCO-$20^i$. Best viewed in color with zoom in.}
\label{figure:visualization}
\end{figure*}

\begin{figure}
\centering
\includegraphics[width=2.6in, height=1.75in]{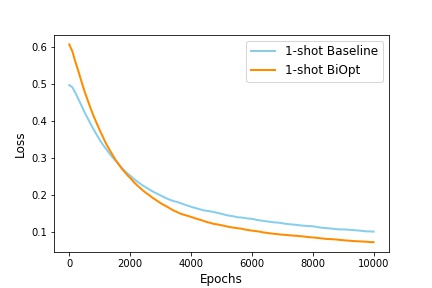}
\caption{Visualization of training loss on PASCAL-$5^i$.}
\label{figure:losscurve}
\end{figure}

\subsection{Visualization}

\textbf{Training Loss}
We visualize training loss of baseline and our BiOpt in Fig. \ref{figure:losscurve}. Orange curve denotes the convergence of our 1-shot model which is optimized through nested loops. Training loss of the base model which is optimized without inner optimization is colored in light blue. It reflects the advantage of the proposed algorithm of bi-level optimization in speeding up convergence. Moreover, training loss of our model is lower after convergence. Our model is supposed to achieve better segmentation performance which is consistent with the conclusion drawn from above Table \ref{tb:innerablation}.

\textbf{Qualitative Visualization}
We display some prediction results in Fig. \ref{figure:visualization}. Our model gives precise segmentation in diverse scenarios. When objects on the support and query images have large appearance and scale variance, clearly seen in the first and last episodes on COCO-$20^i$, BiOpt is able to precisely segment the target object on the query image. Foreground and background objects are separated by our model in these hard episodes. BiOpt is capable of segmenting continuous target regions, such as the \textit{bicycle} on PASCAL-$5^i$.
It demonstrates that our model is capable of estimating representative prototypes by leveraging semantic information on the query image.

\section{Conclusion}
In this paper, we propose the Bi-level Optimization (BiOpt) algorithm for few-shot segmentation. Learning objective is decomposed into two levels and our model is optimized in nested loops. In the inner loop, the model is optimized to learn representative prototypes on the unlabeled query image in each task. In the outer loop, the model is optimized to learn a discriminative embedding space across different tasks.
The advantage of BiOpt is enabling a model capable of leveraging semantic knowledge on the query image into prototype computation in inductive setting. 
We conduct extensive experiments on two benchmarks, furthermore, our BiOpt consistently achieves new state-of-the-art performance in few-shot segmentation.

{\small
\bibliographystyle{ieee_fullname}
\bibliography{cvpr}

\begin{thebibliography}{10}\itemsep=-1pt

\bibitem{chen14semantic}
Liang-Chieh Chen, George Papandreou, Iasonas Kokkinos, Kevin Murphy, and Alan~L
  Yuille.
\newblock Semantic image segmentation with deep convolutional nets and fully
  connected crfs.
\newblock In {\em ICLR}, 2015.

\bibitem{chen2017deeplab}
Liang-Chieh Chen, George Papandreou, Iasonas Kokkinos, Kevin Murphy, and Alan~L
  Yuille.
\newblock Deeplab: Semantic image segmentation with deep convolutional nets,
  atrous convolution, and fully connected crfs.
\newblock {\em PAMI}, 40(4):834--848, 2017.

\bibitem{chen2018encoder}
Liang-Chieh Chen, Yukun Zhu, George Papandreou, Florian Schroff, and Hartwig
  Adam.
\newblock Encoder-decoder with atrous separable convolution for semantic image
  segmentation.
\newblock In {\em ECCV}, pages 801--818, 2018.

\bibitem{cheng2020panoptic}
Bowen Cheng, Maxwell~D Collins, Yukun Zhu, Ting Liu, Thomas~S Huang, Hartwig
  Adam, and Liang-Chieh Chen.
\newblock Panoptic-deeplab: A simple, strong, and fast baseline for bottom-up
  panoptic segmentation.
\newblock In {\em CVPR}, pages 12475--12485, 2020.

\bibitem{dong2018few}
Nanqing Dong and Eric~P Xing.
\newblock Few-shot semantic segmentation with prototype learning.
\newblock In {\em BMVC}, volume~3, 2018.

\bibitem{everingham2010the}
Mark {Everingham}, Luc {Gool}, Christopher~K. {Williams}, John {Winn}, and
  Andrew {Zisserman}.
\newblock The pascal visual object classes (voc) challenge.
\newblock {\em IJCV}, 88(2):303--338, 2010.

\bibitem{finn2017model}
Chelsea Finn, Pieter Abbeel, and Sergey Levine.
\newblock Model-agnostic meta-learning for fast adaptation of deep networks.
\newblock In {\em ICML}, 2017.

\bibitem{he2016deep}
Kaiming {He}, Xiangyu {Zhang}, Shaoqing {Ren}, and Jian {Sun}.
\newblock Deep residual learning for image recognition.
\newblock In {\em CVPR}, pages 770--778, 2016.

\bibitem{hu2019attention}
Tao {Hu}, Pengwan {Yang}, Chiliang {Zhang}, Gang {Yu}, Yadong {Mu}, and Cees
  {Snoek}.
\newblock Attention-based multi-context guiding for few-shot semantic
  segmentation.
\newblock {\em AAAI}, 33:8441--8448, 2019.

\bibitem{kirillov2019panoptic}
Alexander Kirillov, Kaiming He, Ross Girshick, Carsten Rother, and Piotr
  Doll{\'a}r.
\newblock Panoptic segmentation.
\newblock In {\em CVPR}, pages 9404--9413, 2019.

\bibitem{lee2019meta}
Kwonjoon Lee, Subhransu Maji, Avinash Ravichandran, and Stefano Soatto.
\newblock Meta-learning with differentiable convex optimization.
\newblock In {\em CVPR}, 2019.

\bibitem{lin2014microsoft}
Tsung-Yi {Lin}, Michael {Maire}, Serge {Belongie}, Lubomir {Bourdev}, Ross
  {Girshick}, James {Hays}, Pietro {Perona}, Deva {Ramanan}, C.~Lawrence
  {Zitnick}, and Piotr {Dollár}.
\newblock Microsoft coco: Common objects in context.
\newblock {\em arXiv preprint arXiv:1405.0312}, 2014.

\bibitem{liu2020part}
Yongfei {Liu}, Xiangyi {Zhang}, Songyang {Zhang}, and Xuming {He}.
\newblock Part-aware prototype network for few-shot semantic segmentation.
\newblock In {\em ECCV}, 2020.

\bibitem{long2015fully}
Jonathan Long, Evan Shelhamer, and Trevor Darrell.
\newblock Fully convolutional networks for semantic segmentation.
\newblock In {\em CVPR}, pages 3431--3440, 2015.

\bibitem{nguyen2019feature}
Khoi {Nguyen} and Sinisa {Todorovic}.
\newblock Feature weighting and boosting for few-shot segmentation.
\newblock In {\em ICCV}, pages 622--631, 2019.

\bibitem{nichol2018first}
Alex Nichol, Joshua Achiam, and John Schulman.
\newblock On first-order meta-learning algorithms.
\newblock {\em arXiv preprint arXiv:1803.02999}, 2018.

\bibitem{rakelly2018conditional}
Kate {Rakelly}, Evan {Shelhamer}, Trevor {Darrell}, Alyosha~A. {Efros}, and
  Sergey {Levine}.
\newblock Conditional networks for few-shot semantic segmentation.
\newblock In {\em ICLR Workshop}, 2018.

\bibitem{ronneberger2015u}
Olaf Ronneberger, Philipp Fischer, and Thomas Brox.
\newblock U-net: Convolutional networks for biomedical image segmentation.
\newblock In {\em MICCAI}, pages 234--241. Springer, 2015.

\bibitem{russakovsky2015imagenet}
Olga {Russakovsky}, Jia {Deng}, Hao {Su}, Jonathan {Krause}, Sanjeev
  {Satheesh}, Sean {Ma}, Zhiheng {Huang}, Andrej {Karpathy}, Aditya {Khosla},
  Michael {Bernstein}, Alexander~C. {Berg}, and Li {Fei-Fei}.
\newblock Imagenet large scale visual recognition challenge.
\newblock {\em IJCV}, 115(3):211--252, 2015.

\bibitem{shaban2017one}
Amirreza Shaban, Shray Bansal, Zhen Liu, Irfan Essa, and Byron Boots.
\newblock One-shot learning for semantic segmentation.
\newblock 2017.

\bibitem{simonyan2015very}
Karen {Simonyan} and Andrew {Zisserman}.
\newblock Very deep convolutional networks for large-scale image recognition.
\newblock In {\em ICLR}, 2015.

\bibitem{snell2017prototypical}
Jake Snell, Kevin Swersky, and Richard Zemel.
\newblock Prototypical networks for few-shot learning.
\newblock In {\em NIPS}, pages 4077--4087, 2017.

\bibitem{sung2018learning}
Flood Sung, Yongxin Yang, Li Zhang, Tao Xiang, Philip~HS Torr, and Timothy~M
  Hospedales.
\newblock Learning to compare: Relation network for few-shot learning.
\newblock In {\em CVPR}, pages 1199--1208, 2018.

\bibitem{vinyals2016matching}
Oriol Vinyals, Charles Blundell, Timothy Lillicrap, Daan Wierstra, et~al.
\newblock Matching networks for one shot learning.
\newblock In {\em NIPS}, pages 3630--3638, 2016.

\bibitem{wang2020few}
Haochen Wang, Xudong Zhang, Yutao Hu, Yandan Yang, Xianbin Cao, and Xiantong
  Zhen.
\newblock Few-shot semantic segmentation with democratic attention networks.
\newblock ECCV, 2020.

\bibitem{Wang_2019_ICCV}
Kaixin Wang, Jun~Hao Liew, Yingtian Zou, Daquan Zhou, and Jiashi Feng.
\newblock Panet: Few-shot image semantic segmentation with prototype alignment.
\newblock In {\em ICCV}, October 2019.

\bibitem{PMMs2020}
Boyu Yang, Chang Liu, Bohao Li, Jianbin Jiao, and Qixiang Ye.
\newblock Prototype mixture models for few-shot semantic segmentation.
\newblock In {\em ECCV}, 2020.

\bibitem{zhang2019pyramid}
Chi Zhang, Guosheng Lin, Fayao Liu, Jiushuang Guo, Qingyao Wu, and Rui Yao.
\newblock Pyramid graph networks with connection attentions for region-based
  one-shot semantic segmentation.
\newblock In {\em ICCV}, pages 9587--9595, 2019.

\bibitem{zhang2019canet}
Chi Zhang, Guosheng Lin, Fayao Liu, Rui Yao, and Chunhua Shen.
\newblock Canet: Class-agnostic segmentation networks with iterative refinement
  and attentive few-shot learning.
\newblock In {\em CVPR}, pages 5217--5226, 2019.

\bibitem{zhang2020sg}
Xiaolin {Zhang}, Yunchao {Wei}, Yi {Yang}, and Thomas {Huang}.
\newblock Sg-one: Similarity guidance network for one-shot semantic
  segmentation.
\newblock {\em IEEE Transactions on Cybernetics}, 2020.

\end{thebibliography}
}

\end{document}